\documentclass[10pt,twocolumn,letterpaper]{article}

\usepackage{iccv}
\usepackage{times}
\usepackage{epsfig}
\usepackage{graphicx}
\usepackage{amsmath}
\usepackage{amssymb}

\usepackage[sectionbib,numbers,sort]{natbib}
\usepackage{amsmath}
\usepackage{subcaption}
\usepackage{soul}
\usepackage{rotating}
\usepackage{float}

\usepackage[ruled,vlined,linesnumbered]{algorithm2e}

\usepackage{graphicx}
\usepackage[export]{adjustbox}
\usepackage{xcolor}

\usepackage{array}
\usepackage{booktabs}
\usepackage{multirow}
\usepackage{pifont}

\newcolumntype{f}{>{$}l<{$}}
\newcolumntype{n}{l}
\newcolumntype{N}{>{\scriptsize}l}
\newcolumntype{v}[1]{>{\raggedright\hspace{0pt}}p{#1}}
\newcolumntype{V}[1]{>{\scriptsize\raggedright\hspace{0pt}}p{#1}}
\newcolumntype{C}[1]{>{\scriptsize\centering\hspace{0pt}}p{#1}}
\newcolumntype{L}[1]{>{\scriptsize\hspace{0pt}}p{#1}}
\newcolumntype{B}[1]{>{\boldmath\DC@{.}{,}{#1}}l<{\DC@end}}
\newcolumntype{d}[1]{>{\DC@{.}{,}{#1}}l<{\DC@end}}
\newcolumntype{R}[1]{%
>{\begin{turn}{90}\begin{minipage}{#1}\scriptsize\raggedright\hspace{0pt}}l%
<{\end{minipage}\end{turn}}%
}
\newcolumntype{x}{>{\scriptsize\raggedright\hspace{0pt}}X}

\renewcommand{\vec}[1]{\mathbf{\boldsymbol{#1}}}

\newcommand{\predTh}[1]{{\operatorname{\mathsf{\color{blue!96!black}#1}}}}
\newcommand{\predThF}[1]{{\operatorname{\mathsf{#1}}}}

\newcommand{\timePoints}[0]{$\mathcal{\color{blue!96!black}T}$}

\newcommand{\actionsEvents}[0]{${\color{blue!96!black}\Theta}$}

\newcommand{\objectSort}[0]{$\mathcal{\color{blue!96!black}O}$}

\newcommand{\bul}{{\small$\color{blue!80!black}\blacktriangleright~$}}

\newcommand{\bulletpoint}[1]{\null\quad -- $~$ \begin{minipage}[t]{0.9\columnwidth}{#1}\end{minipage}\\[2pt]}

\definecolor{YellowGreen}{RGB}{160,200,40}
\definecolor{mathcolor}{RGB}{7,72,110}

\usepackage[pagebackref=true,breaklinks=true,letterpaper=true,colorlinks,bookmarks=false]{hyperref}

\hypersetup{
    pdftitle={Commonsense Scene Semantics for Cognitive Robotics }, 
    pdfauthor={Jakob Suchan, Mehul Bhatt},     	
    pdfsubject={Commonsense Scene Semantics for Cognitive Robotics}, 
    pdfkeywords={Cognitive Robotics} {Visuo-Locomotive Interactions} {Spatio-Temporal Dynamics}, 					
    unicode=false,          									
    pdftoolbar=false,        									
    pdfmenubar=true,        								
    pdffitwindow=false,     									
    pdfstartview={FitH},    								
    pdfnewwindow=true,      								
    bookmarks=true,         								
    colorlinks=true,       									
    linkcolor=red!70!black,          							
    citecolor=blue!90!black,        							
    filecolor=blue,      									
    urlcolor=blue!60!black          								
}

 \iccvfinalcopy 

\def\iccvPaperID{****} 

\ificcvfinal\fi
\begin{document}


\title{\sffamily Commonsense Scene Semantics for Cognitive Robotics \\[6pt] \textnormal{\normalsize\sffamily Towards Grounding Embodied Visuo-Locomotive Interactions}}






\author{{\small\sffamily Jakob Suchan$^1$, and Mehul Bhatt$^{1,2}$}\\[6pt] 
\sffamily\small$^1$Spatial Reasoning. \url{www.spatial-reasoning.com}\\[2pt]\sffamily\small EASE CRC: Everyday Activity Science and Engineering., \url{http://ease-crc.org}\\\sffamily\small University of Bremen, Germany\\\sffamily\small$^2$Machine Perception and Interaction Lab., \url{https://mpi.aass.oru.se}\\\sffamily\small Centre for Applied Autonomous Sensor Systems (AASS)\\\sffamily\small \"{O}rebro University, Sweden
}

\maketitle
\thispagestyle{empty}

\begin{abstract}
{\small\sffamily\upshape{We present a commonsense, qualitative model for the semantic grounding of embodied visuo-spatial and locomotive interactions. The key contribution is an integrative methodology combining low-level visual processing with high-level, human-centred representations of space and motion rooted in artificial intelligence. We demonstrate practical applicability with examples involving object interactions, and indoor movement.}
}
\end{abstract}

\section{Introduction}
Practical robotic technologies and autonomous systems in real-world settings are confronted with a range of situational and context-dependent challenges from the viewpoints of perception \& sensemaking, high-level planning \& decision-making, human-machine interaction etc. Very many research communities and sub-fields thereof addressing the range of challenges from different perpecsectives have flourished in the recent years: computer vision, artificial intelligence, cognitive systems, human-machine interaction, cognitive science, multi-agent systems, control \& systems engineering to name a few.  Driven by the need to achieve contextualised practical deployability in real-world non-mundane everyday situations involving living beings, there is now a clearly recognised need for integrative research that combines state of the art methods from these respective research areas. Towards this, the research presented in this paper addresses commonsense visuo-spatial scene interpretation in indoor robotics settings at the interface of vision, AI, and spatial cognition. The focus of this research is on activities of everyday living involving people, robots, movement, and human-machine interaction.


\medskip

\textbf{Interpreting Embodied Interaction:\\[2pt]On Grounded Visuo-Locomotive Perception}

Visuo-locomotive perception denotes the capability to develop a conceptual mental model (e.g., consisting of abstract, commonsense representations) emanating from multi-sensory perceptions during embodied interactions and movement in a real world populated by static and dynamic entities and artefacts (e.g., moving objects, furniture). Visuo-locomotive perception in the context of cognitive robotics technologies and machine perception \& interaction systems involves a complex interplay of high-level cognitive processes. These could, for instance, encompass capabilities such as explainable reasoning, learning, concept formation, sensory-motor control; from a technical standpoint of AI technologies, this requires the mediation of commonsense formalisms for reasoning about \emph{space, events, actions, change}, and \emph{interaction} \citep{Bhatt:RSAC:2012}. 

\smallskip

\bul $~$ With visuo-locomotive cognition as the context, consider the task of semantic interpretation of multi-modal perceptual data (e.g., about human behaviour, the environment and its affordances), with objectives ranging from knowledge acquisition and data analyses to hypothesis formation, structured relational learning, learning by demonstration etc. Our research focusses on the processing and semantic interpretation of dynamic visuo-spatial imagery with a particular emphasis on the ability to abstract, reason, and learn commonsense knowledge that is semantically founded in qualitative spatial, temporal, and spatio-temporal relations and motion patterns. We propose that an ontological characterisation of human-activities --- e.g., encompassing (embodied) spatio-temporal relations--- serves as a bridge between high-level conceptual categories (e.g., pertaining to human-object interactions) on the one-hand, and low-level / quantitative sensory-motor data on the other.

\begin{figure*}[t]
    \centering
    \includegraphics[width=0.95\textwidth]{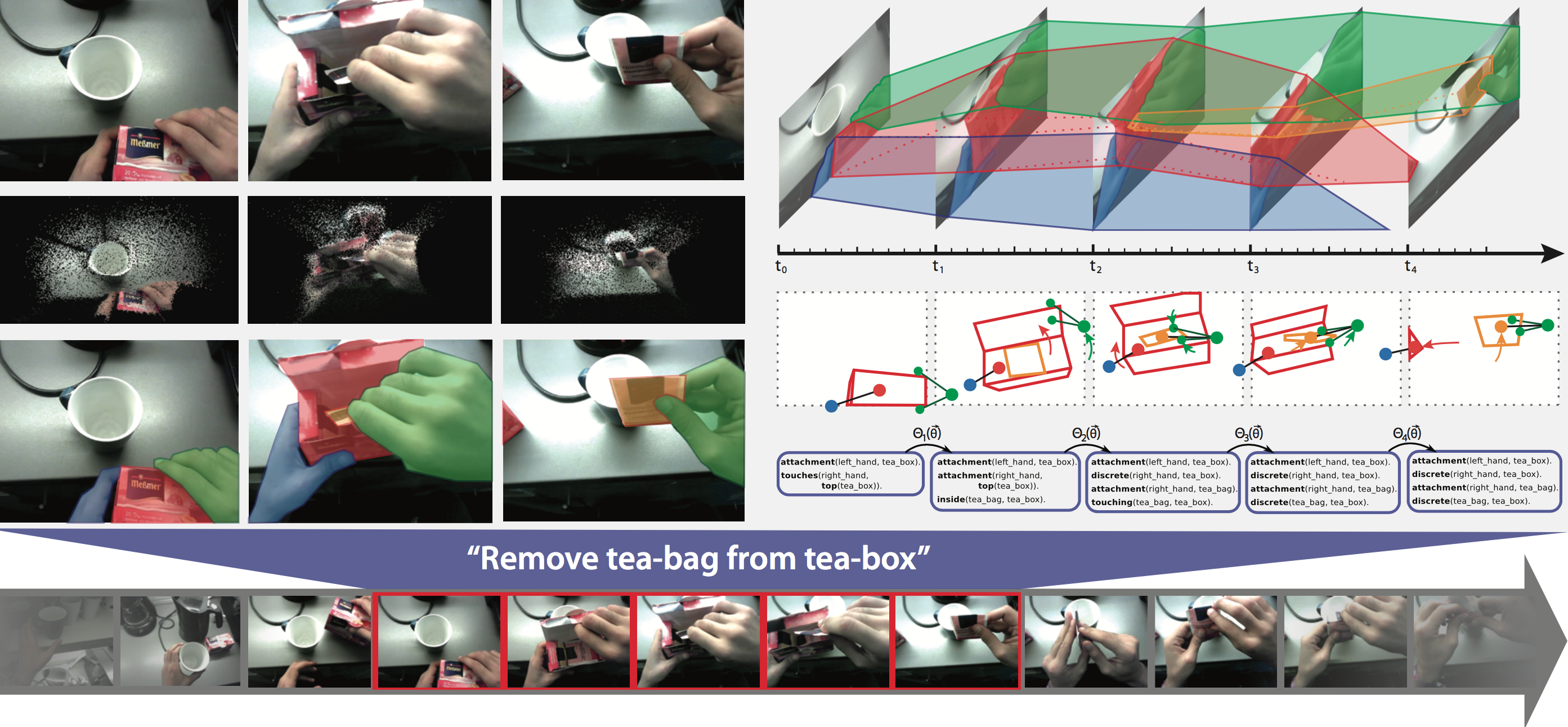}
    \caption{{\sffamily\small A Sample Activity -- ``Making a cup of tea'' (egocentric view from a head-mounted RGB-D capture device)}}
    \label{fig:tea-box}
\end{figure*}

\medskip

\textbf{Commonsense Scene Semantics:\\[2pt]Integrated Vision and Knowledge Representation}

The starting point of the work presented in this paper is in formal commonsense representation and reasoning techniques developed in the field of Artificial Intelligence. Here, we address the key question:

\smallskip

{\sffamily\small How can everyday embodied activities (involving interaction and movement) be formally represented in terms of spatio-temporal relations and movement patterns (augmented by context-dependent knowledge about objects and environments) such that the representation enables robotic agents to execute everyday interaction tasks  (involving manipulation and movement) appropriately?}

\smallskip

We particularly focus on an ontological and formal characterisation of space and motion from a human-centered, commonsense formal modeling and computational reasoning viewpoint,
i.e., \emph{space-time}, as it is interpreted within the AI subdisciplines of knowledge representation and reasoning, and commonsense reasoning, and within spatial cognition \& computation,
and more broadly, within spatial information theory \citep{Cohn-Renz-07-QSRKRHandbook,renz-nebel-hdbk07,hdbook-spatial-logics,bhatt2011-scc-trends,Bhatt:RSAC:2012,Bhatt-Schultz-Freksa:2013}.

\smallskip

\bul $~$ We build on state of the art methods for visual processing of RGB-D and point-cloud data for sensing the environment and the people within. In focus are 3D-SLAM data for extracting floor-plan structure based on plane detection in point-clouds, and people detection and skeleton tracking using {\footnotesize\sffamily Microsoft Kinect v2}. Furthermore, we combine robot self-localisation and people tracking to localise observed people interactions in the global space of the environmental map.

\medskip

We emphasise that the ontological and representational aspects of our research are strongly driven by computational considerations focussing on:\quad \textbf{\small(a)}.  developing general methods and tools for commonsense reasoning about space and motion categorically from the viewpoint of commonsense cognitive robotics in general, but human-object interactions occurring in the context of everyday activities in particular;\quad \textbf{\small(b)}. founded on the established ontological model, developing models, algorithms and tools for reasoning about space and motion, and making them available as part of cognitive robotics platforms and architectures such as ROS. The running examples presented in the paper highlight the semantic question-answering capabilities that are directly possible based on our commonsense model directly in the context of \emph{constraint logic programming}.


\smallskip
%


\begin{table*}[t]
\centering
\scriptsize\sffamily
\begin{tabular}{|l|p{15 ex}|p{45.2 ex}|p{22 ex}|}
\hline
\textbf{\color{blue!70!black}\textsc{Spatial Domain} ($\mathcal{QS}$)} & \emph{Formalisms}  & \emph{Spatial Relations ({\color{blue!70!black}$\mathcal{R}$})} & \emph{Entities ({\color{blue!90!black}$\mathcal{E}$})} \\
\hline
\hline
\multirow{2}{*}{{\color{blue!70!black}Mereotopology}} & {\tiny RCC-5, RCC-8 \citep{randell1992spatial}} & {\scriptsize\sffamily disconnected (dc), external contact (ec), partial overlap (po), tangential proper part (tpp), non-tangential proper part (ntpp), proper part (pp), part of (p), discrete (dr), overlap (o), contact (c)} & arbitrary rectangles, circles, polygons, cuboids, spheres \\
\cline{2-4}
&  \tiny Rectangle \& Block algebra \citep{guesgen1989spatial} & {\scriptsize\sffamily proceeds, meets, overlaps, starts, during, finishes, equals} & axis-aligned rectangles and cuboids \\
 \hline
\multirow{2}{*}{{\color{blue!70!black}Orientation}}  & \tiny LR \citep{Scivos2004} & {\scriptsize\sffamily left, right, collinear, front, back, on} & 2D point, circle, polygon with 2D line\\
\cline{2-4}
& \tiny OPRA \citep{moratz06_opra-ecai} & {\scriptsize\sffamily facing towards, facing away, same direction, opposite direction} & oriented points, 2D/3D vectors \\
 \hline
\multirow{2}{*}{{\color{blue!70!black}Distance, Size}}   & \tiny QDC \citep{hernandez1995qualitative} & {\scriptsize\sffamily adjacent, near, far, smaller, equi-sized, larger} & rectangles, circles, polygons, cuboids, spheres\\
 \hline
\multirow{2}{*}{{\color{blue!70!black}Dynamics, Motion}}   & \tiny Space-Time Histories \cite{Hayes:Naive-I, Hazarika:2005:thesis} & {\scriptsize\sffamily moving: towards, away, parallel; growing / shrinking:  vertically, horizontally; passing: in front, behind; splitting / merging; rotation: left, right, up, down, clockwise, couter-clockwise} & rectangles, circles, polygons, cuboids, spheres\\
\hline
\end{tabular}

$~$
  
  \caption{{\small\sffamily Commonsense Spatio-Temporal Relations for Abstracting Space and Motion in Everyday Human Interaction}}
\label{tbl:relations}
\end{table*}

\section{Commonsense, Space, Motion} 
Commonsense spatio-temporal relations and patterns (e.g. {\sffamily\small {left-of}, {touching}, {part-of}, {during}, {approaching, collision}}) offer a human-centered and cognitively adequate formalism for logic-based automated reasoning about embodied spatio-temporal interactions involved in everyday activities
such as \emph{flipping a pancake}, \emph{grasping a cup}, or \emph{opening a tea box} \citep{Bhatt-Schultz-Freksa:2013, worgotter2012simple,PRICAI-2014-spatial,Bhatt-2016-IJCAI-NLP}. Consider Fig. \ref{fig:tea-box}, consisting of a sample human activity ---``\emph{making a cup of tea}''--- as captured from an egocentric viewpoint with a head-mounted RGB-D capture device. From a commonsense viewpoint, the sequence of high-level steps typically involved in this activity, e.g.,  opening a tea-box, removing a tea-bag from the box and putting the tea-bag inside a tea-cup filled with water while holding the tea-cup, each qualitatively correspond to high-level spatial and temporal relationships between the agent and other involved objects. For instance, one may most easily identify relationships of {\sffamily\small {contact} and {containment}} that hold across specific time-intervals. Here, parametrised manipulation or control actions ({\small $\Theta_{1}(\vec{\theta})$, ...$\Theta_{n}(\vec{\theta})$}) effectuate state transitions, which may be qualitatively modelled as changes in topological relationships amongst involved domain entities.

\smallskip

Embodied interactions, such as those involved in Fig. \ref{fig:tea-box}, may be grounded using a holistic model for the commonsense , qualitative representation of space, time, and motion (Table \ref{tbl:relations}). In general, qualitative, multi-modal, multi-domain\footnote{Multi-modal in this context refers to more than one aspect of space, e.g., topology, orientation, direction, distance, shape. Multi-domain denotes a mixed domain ontology involving points, line-segments, polygons, and regions of space, time, and space-time \citep{Hazarika:2005:thesis}. Refer Table \ref{tbl:relations}.} 
 representations of spatial, temporal, and spatio-temporal relations and patterns, and their mutual transitions can provide a mapping and mediating level between human-understandable natural language instructions and formal narrative semantics on the one hand \citep{CR-2013-Narra-CogRob,Bhatt-Schultz-Freksa:2013}, and symbol grounding, quantitative trajectories, and low-level primitives for robot motion control on the other. By spatio-linguistically grounding complex sensory-motor trajectory data (e.g., from human-behaviour studies) to a formal framework of space and motion, generalized (activity-based) qualitative reasoning about dynamic scenes, spatial relations, and motion trajectories denoting single
and multi-object path \& motion predicates can be supported \citep{Eschenbach-Schill-99}. For instance, such predicates can be abstracted within a region based 4D space-time framework \citep{Hazarika:2005:thesis,DBLP:conf/aaai/BennettCTH00,DBLP:conf/ecai/BennettCTH00}, object interactions \citep{DBLP:journals/ai/Davis08,DBLP:journals/ai/Davis11}, and spatio-temporal narrative knowledge \citep{TylerEvans2003,CR-2013-Narra-CogRob,DBLP:journals/scc/Davis13}.
An adequate qualitative spatio-temporal representation can therefore connect with low-level constraint-based movement
control systems of robots \citep{bartels13constraints}, and also help grounding symbolic descriptions of actions and objects to be manipulated (e.g., natural language instructions such as cooking recipes \citep{tellex2010natural}) in the robots perception.

\section{Visuo-Locomotive Interactions:\\A Commonsense Characterisation}
\label{sec:activitiy_abstractions}



%

\subsection{Objects and Interactions in Space-Time} 

Activities and interactions are described based on visuo-spatial domain-objects \objectSort\ = $\{o_1, o_2, ... , o_i\}$ representing the visual elements in the scene, e.g., \emph{people} and {objects}. The \textbf{Qualitative Spatio-Temporal Ontology} ({\color{blue}$\mathcal{QS}$}) is characterised by the basic spatial and temporal  entities ({\color{blue}$\mathcal{E}$}) that can be used as abstract representations of domain-objects and the relational spatio-temporal structure ({\color{blue}$\mathcal{R}$}) that characterises the qualitative spatio-temporal relationships amongst the entities in ({\color{blue}$\mathcal{E}$}). 
%
Towards this, domain-objects (\objectSort) are represented by their spatial and temporal properties, and abstracted using the following basic \emph{spatial entities}:\\[4pt]
%
%
{\small
\bulletpoint{\emph{points} are triplets of reals $x,y,z$;}
\bulletpoint{\emph{oriented-points}  consisting of a point $p$ and a vector $v$; }
\bulletpoint{\emph{line-segments}  consisting of two points $p_1, p_2$ denoting the start and the end point of the line-segment; }
\bulletpoint{\emph{poly-line} consisting of a list of vertices (points) $p_1$, ..., $p_n$, such that the line is connecting the vertices is non-self-intersecting; }
\bulletpoint{\emph{polygon} consisting of a list of vertices (points) $p_1$, ..., $p_n$, (spatially ordered counter-clockwise) such that the boundary is non-self-intersecting;  }
}

and the temporal entities: \\[4pt]
{\small
\bulletpoint{\emph{time-points} are a real $t$}
\bulletpoint{\emph{time-intervals} are a pair of reals $t_1,t_2$, denoting the start and the end point of the interval.}
}


The dynamics of human activities are represented by 4-dimensional regions in space-time ({\sffamily sth}) representing people and object dynamics by a set of spatial entities in time, i.e. $\color{blue}\mathcal{STH}$ = ($\varepsilon_{t_1}, \varepsilon_{t_2}, \varepsilon_{t_3}, ..., \varepsilon_{t_n}$), where $\varepsilon_{t_1}$ to $\varepsilon_{t_n}$ denotes the spatial primitive representing the object $o$ at the time points $t_1$ to $t_n $.

 \begin{figure}
\begin{minipage}{\columnwidth} 
\centering
\includegraphics[height = 2.5in]{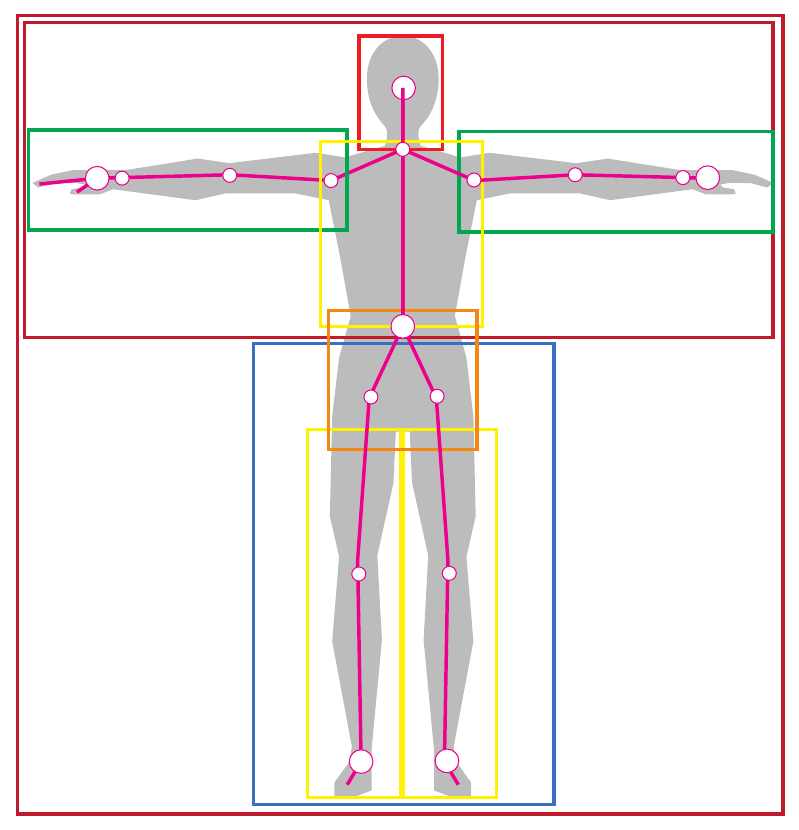}
\end{minipage}
\begin{minipage}{\columnwidth} 
\scriptsize
\includegraphics[width = 0.95\columnwidth]{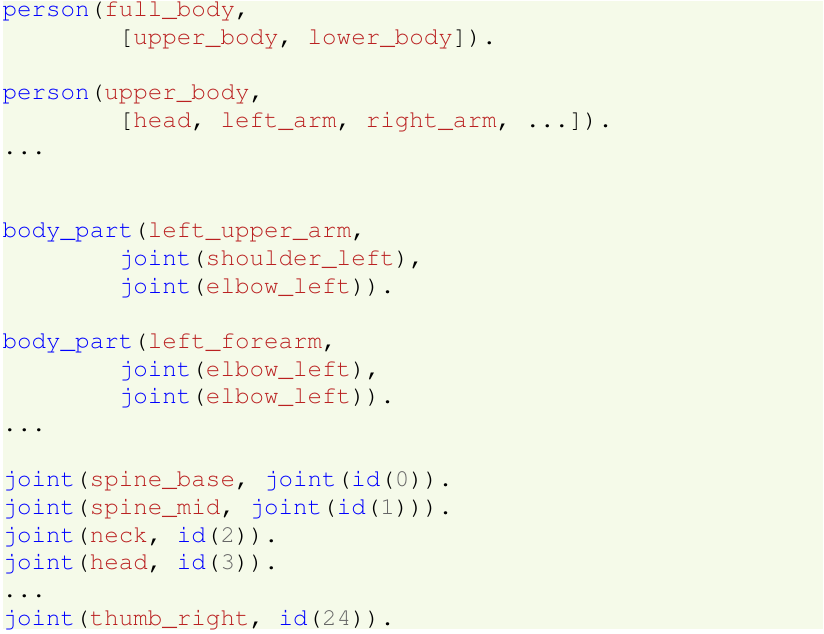}

%
%
%
%
%
%
\end{minipage}
\caption{{\small\sffamily Declarative Model of Human-Body Posture}}
\label{fig:human_body}
\end{figure}

\medskip
\medskip
\medskip
\medskip

\textbf{Declarative Model of Visuo-Locomotive Interactions}\quad
Based on the qualitative spatio-temporal ontology  ({$\mathcal{QS}$}), human interactions in the environment are represented using a declarative model of visuo-locomotive interactions, encompassing dynamics of \emph{humans}, \emph{objects}, and the \emph{environmental characteristics}.   
 

\begin{itemize}
\item \emph{Human Body Pose} \quad
The human body is represented using a declarative model of the body-structure (see Fig. \ref{fig:human_body}), within this model we ground the human body-pose in 3D-data of skeleton joints and body-parts obtained from RGB-D sensing.

\item\emph{Semantics of the Environment}\quad
The semantic structure of the environment is represented using a topological map corresponding to the floor-plan of the environment extracted from 3D point-clouds obtained from 3D-SLAM data.

\end{itemize}

Using these models, visuo-locomotive interactions, involving humans, robots, and objects can be declaratively abstracted by the spatio-temporal characteristics of the involved domain-objects, and may be used for high-level interpretation and reasoning about scene dynamics.

\begin{figure*}[t]
 
 \tiny
 
\centering
 
\subcaptionbox*{\scriptsize$\predTh{discrete}(o_1,o_2)$}{\includegraphics[height = 0.72in]{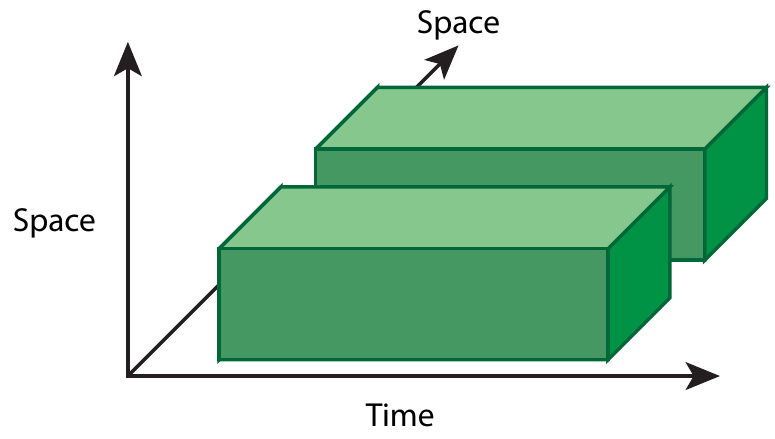}}\quad
\subcaptionbox*{\scriptsize$\predTh{overlapping}(o_1,o_2)$}{\includegraphics[height = 0.72in]{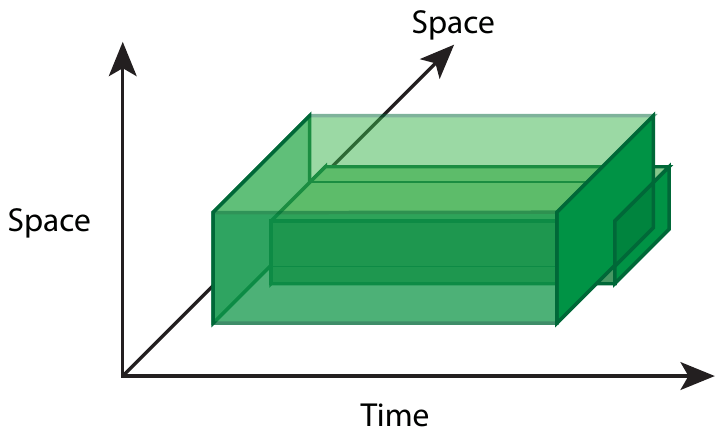}}\quad
\subcaptionbox*{\scriptsize$\predTh{inside}(o_1,o_2)$}{\includegraphics[height = 0.72in]{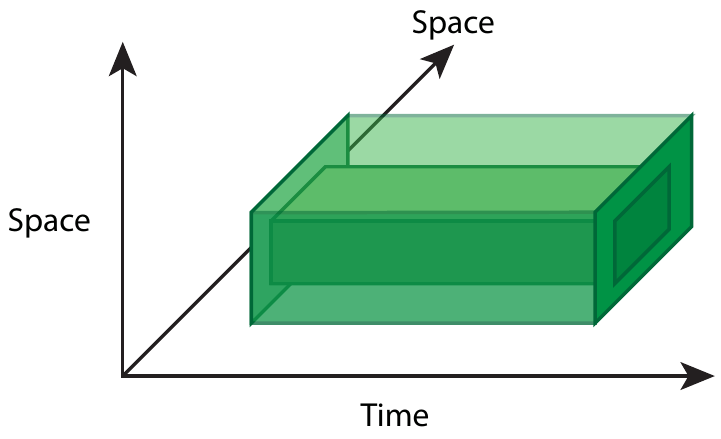}}\quad
\subcaptionbox*{\scriptsize$\predTh{moving}(o)$}{\includegraphics[height = 0.72in]{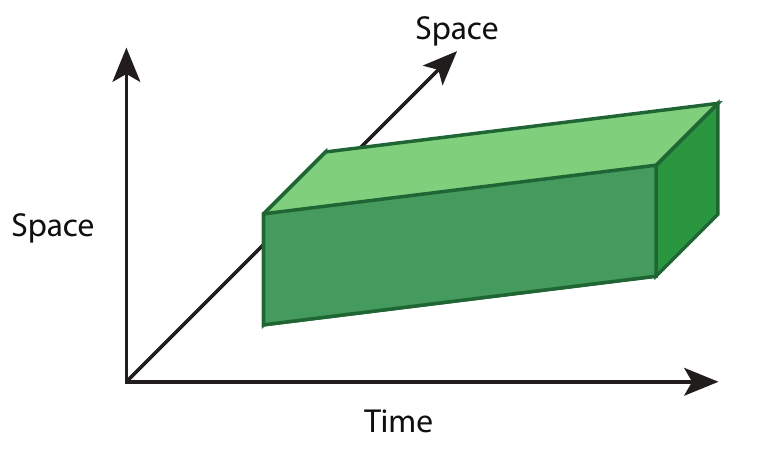}}\quad
\subcaptionbox*{\scriptsize$\predTh{stationary}(o)$}{\includegraphics[height = 0.72in]{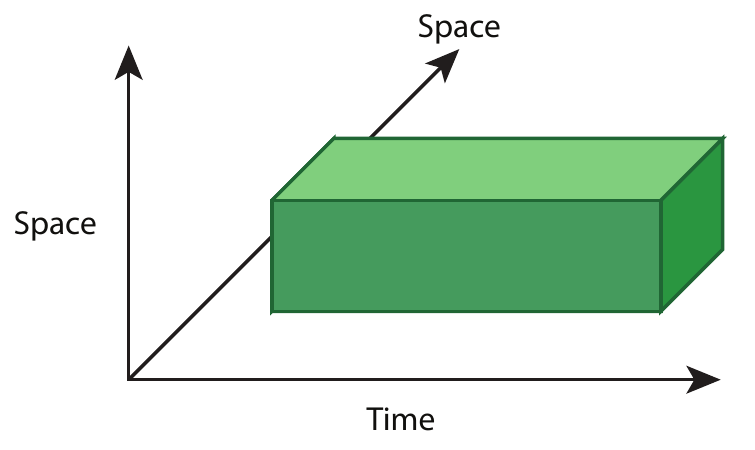}}

\vspace{8pt}
\subcaptionbox*{\scriptsize$\predTh{growing}(o)$}{\includegraphics[height = 0.72in]{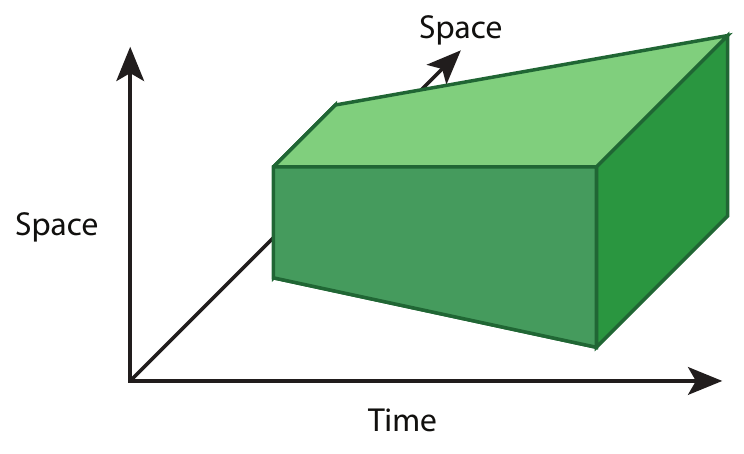}}\quad
\subcaptionbox*{\scriptsize$\predTh{shrinking}(o)$}{\includegraphics[height = 0.72in]{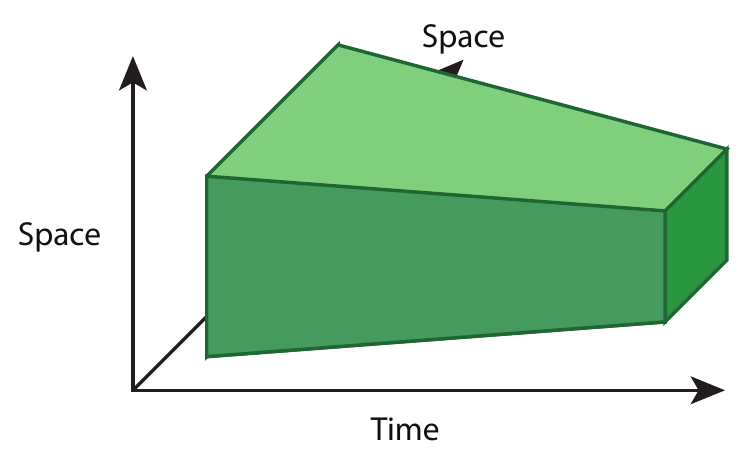}}\quad
\subcaptionbox*{\scriptsize$\predTh{parallel}(o_1,o_2)$}{\includegraphics[height = 0.72in]{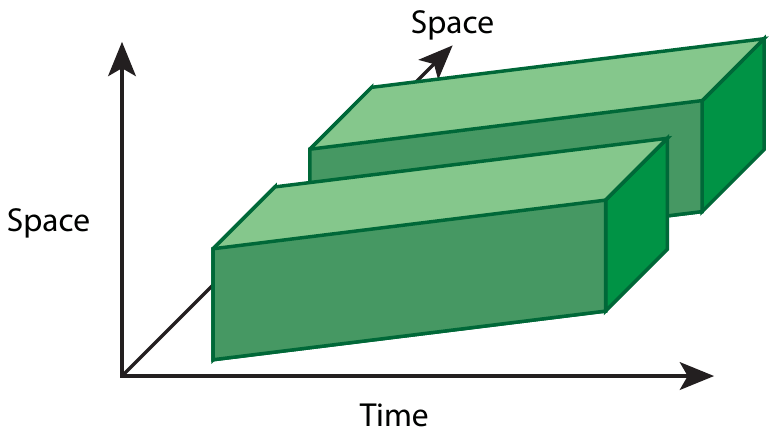}}\quad
\subcaptionbox*{\scriptsize$\predTh{merging}(o_1,o_2)$}{\includegraphics[height = 0.72in]{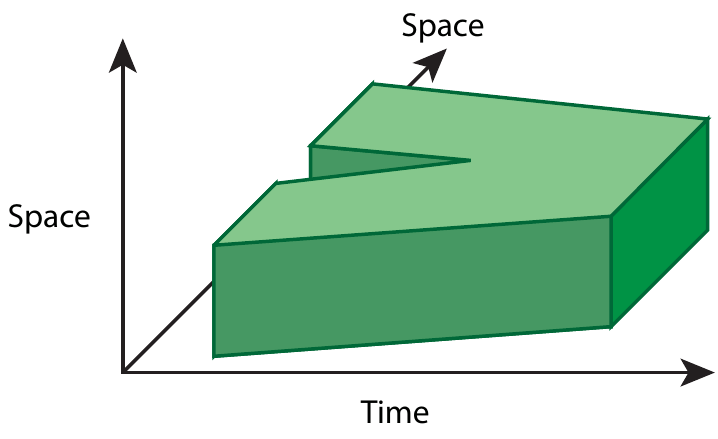}}\quad
\subcaptionbox*{\scriptsize$\predTh{splitting}(o_1,o_2)$}{\includegraphics[height = 0.72in]{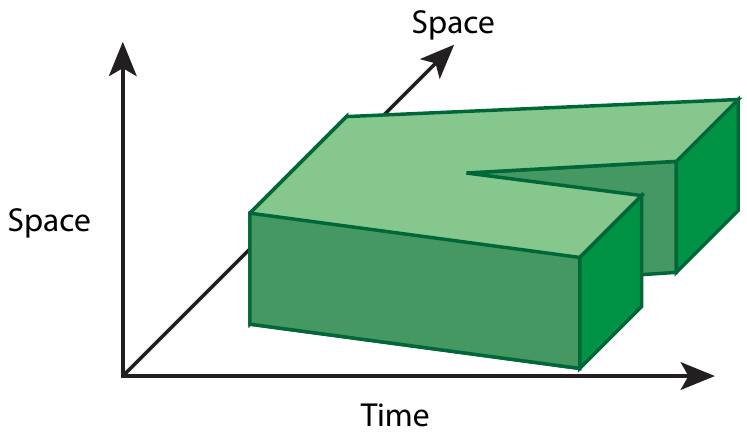}}

\vspace{8pt}

\subcaptionbox*{\scriptsize$\predTh{curved}(o)$}{\includegraphics[height = 0.7in]{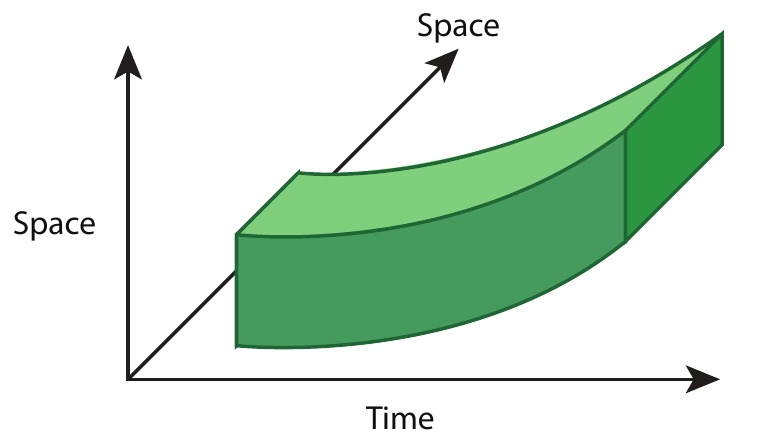}}\quad
\subcaptionbox*{\scriptsize$\predTh{cyclic}(o)$}{\includegraphics[height = 0.72in]{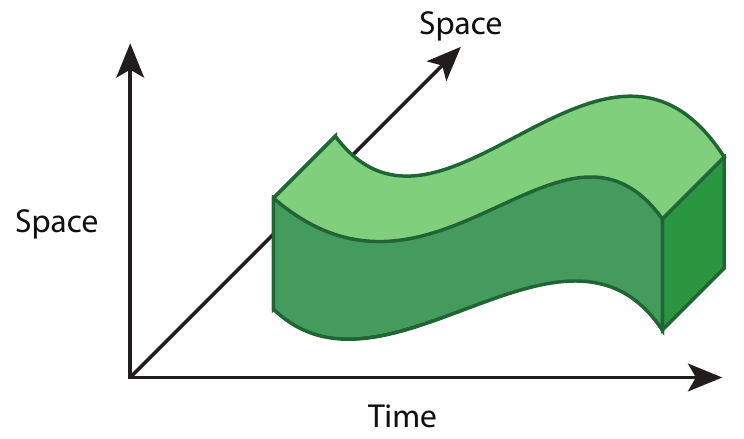}}\quad
\subcaptionbox*{\scriptsize$\predTh{moving\_into}(o_1,o_2)$}{\includegraphics[height = 0.72in]{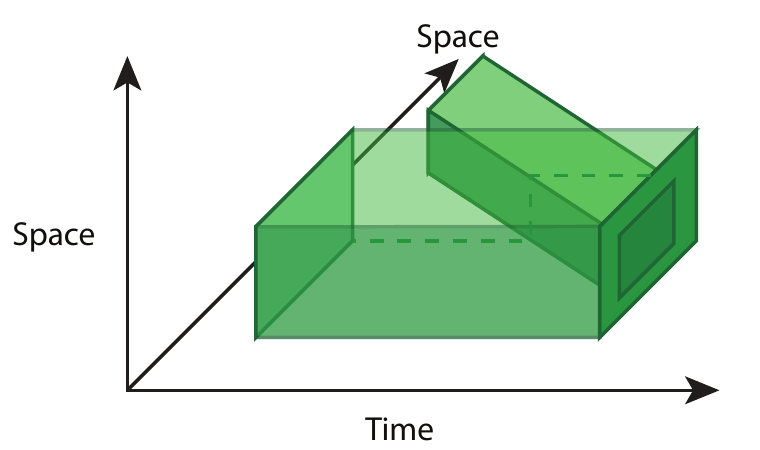}}\quad
\subcaptionbox*{\scriptsize$\predTh{moving\_out}(o_1,o_2)$}{\includegraphics[height = 0.72in]{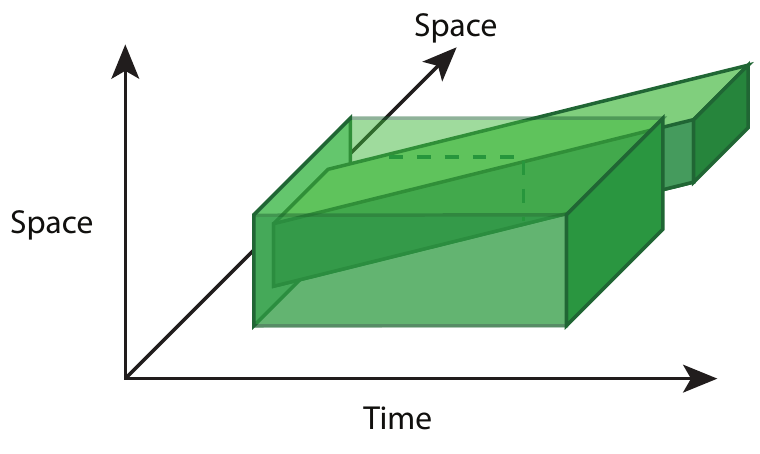}}\quad
\subcaptionbox*{\scriptsize$\predTh{attached}(o_1,o_2)$}{\includegraphics[height = 0.72in]{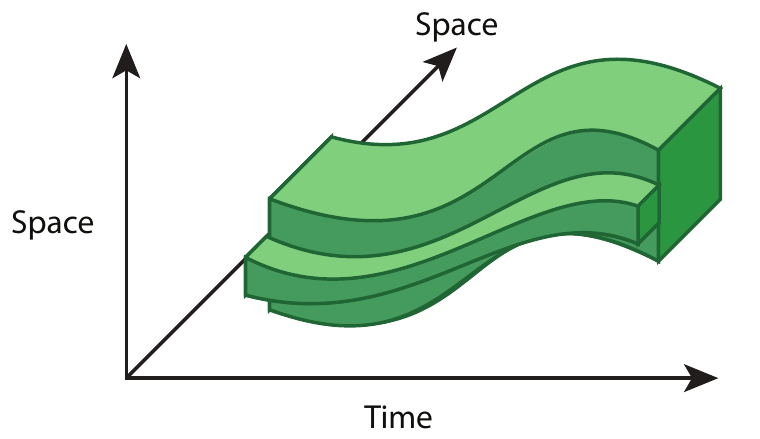}}

\caption{{\small\sffamily Commonsense Spatial Reasoning with Space-Time Histories Representing Dynamics in Everyday Human Activities}}
\label{fig:s-t-entities}
\end{figure*}

\subsection{Spatio-Temporal Characteristics\\of Human Activities} 


The \emph{space-time histories} (sth) used to abstract the dynamics of human activities are based on basic spatio-temporal entities obtained from the sensed data, corresponding to the declarative model of visuo-locomotive interactions.
To extract these entities, we define functions for the specific spatio-temporal properties of domain-objects. I.e., the following functions are used for static spatial properties.


\begin{itemize}
\item[--] position: {\small \objectSort $\times$ \timePoints $\rightarrow \mathbb{R} \times  \mathbb{R} \times \mathbb{R}$}, gives
the 3D position (x,y,z) of an object $o$ at a  time-point $t$;  
\item[--] $size$: {\small\objectSort $\times$ \timePoints $\rightarrow \mathbb{R}$}, gives the size of an object $o$ at a time-point $t$;
\item[--] $distance$:  {\small\objectSort $\times$ \objectSort $\times$ \timePoints $\rightarrow \mathbb{R}$}, gives the distance between two objects $o_1$ and $o_2$ at a time-point $t$;
\item[--] $angle$: {\small\objectSort $\times$ \objectSort $\times$ \timePoints $\rightarrow \mathbb{R}$}, gives the angle between two objects $o_1$ and $o_2$ at a  time-point $t$;
\end{itemize}

To account for changes in the spatial properties of domain-objects we use the following functions for dynamic spatio-temporal properties.

\begin{itemize}
\item[--] $movement\ velocity$: {\small\objectSort $\times$ \timePoints $\times$ \timePoints $\rightarrow \mathbb{R}$}, gives the amount of movement of an object $o$ between two time-points $t_1$ and $t_2$;
\item[--] $movement\ direction$: {\small\objectSort $\times$ \timePoints $\times$ \timePoints $\rightarrow \mathbb{R}$}, gives the direction of movement of an object $o$ between two time-points $t_1$ and $t_2$;
\item[--] $rotation$: {\small\objectSort $\times$ \timePoints $\times$ \timePoints $\rightarrow \mathbb{R}$}, gives the rotation of an object $o$ between two time-points $t_1$ and $t_2$;
\end{itemize}

These functions are used to obtain basic spatial entities, e.g. \emph{points, lines, regions}, from the sensor data.
Spatio-temporal relationships ($\mathcal{R}$) between the basic entities in $\mathcal{E}$ may be characterised with respect to arbitrary spatial and spatio-temporal domains such as \emph{mereotopology, orientation, distance, size, motion, rotation} (see Table \ref{tbl:relations} for a list of considered spatio-temporal abstractions). 
E.g, let $D_1, \dots, D_n$ be spatial domains (e.g. the domain of axis-aligned rectangles). A spatial relation $r$ of arity $n$ ($0 < n$) is defined as:

$$ r \subseteq D_1 \times \dots \times D_n.$$

The spatio-temporal dynamics of the scene can then be represented based on the relations holding between the objects in the scene, and the changes with in them.

\smallskip


\textbf{Spatio-temporal fluents} are used to describe properties of the world, i.e. the predicates ${\color{blue}\predThF{holds-at}}(\phi, t)$  and ${\color{blue}\predThF{holds-in}}(\phi, \delta)$ denote that the fluent $\phi$ holds at time point $t$, resp. in time interval $\delta$. 
Fluents are determined by the data from the depth sensing device and represent qualitative relations between domain-objects, i.e. spatio-temporal fluents denote, that a relation $r \in  \mathcal{R}$ holds between basic spatial entities  $\varepsilon$ of a space-time history at a time-point $t$. Dynamics of the domain are represented based on changes in spatio-temporal fluents (see Fig. \ref{fig:s-t-entities}), e.g., two objects approaching each other can be defined as follows.

\noindent\begin{minipage}{\columnwidth}
{\footnotesize
\begin{align}
\begin{split}
&\predThF{holds-in}(\mathsf{approaching}(o_i, o_j), \delta) \supset \mathsf{during}(t_i, \delta) \wedge \mathsf{during}(t_j, \delta) \wedge \\
& \quad \mathsf{before}(t_i, t_j) \wedge (distance(o_i, o_j, t_i) > distance(o_i, o_j, t_j)).\\
\end{split}
\end{align}}
\end{minipage}

\begin{table}[t]
\begin{center}
\scriptsize\sffamily
\begin{tabular}{|p{2.5cm}|p{4.8cm}|}

\hline
\textbf{Interaction} (\actionsEvents) &  Description \\

\hline

$pick\_up(P, O)$ & a person $P$ picks up an object $O$. \\

$put\_down(P, O)$  & a person $P$ puts down an object $O$. \\

$reach\_for(P, O)$  & a person $P$ is reaches for an object $O$. \\

$pass\_over(P_1, P_2, O)$ & a person $P_1$ passes an object $O$ to another person $P_2$.\\

\hline

\end{tabular}

\medskip

\begin{tabular}{|p{2.5cm}|p{4.8cm}|}

\hline
\textbf{Interaction} (\actionsEvents) &  Description \\

\hline

$moves\_into(P, FS)$ & a person $P$ enters a floor-plan structure $FS$. \\

$passes(P, FS)$  & a person $P$ passes through a floor-plan structure $FS$. \\

\hline

\end{tabular}
\end{center}
\caption{{\small\sffamily Sample Interactions Involved in Everyday Human Activities:  Human Object Interactions and Peoples Locomotive Behaviour}}
\label{tbl:interactions}
\end{table}%

\textbf{Interactions}.\quad Interactions \actionsEvents\ $= \{\theta_1, \theta_2, ... , \theta_i\}$ describe processes that change the spatio-temporal configuration of objects in the scene, at a specific time; these are defined by the involved spatio-temporal dynamics in terms of  changes in the status of space-time histories caused by the interaction, i.e. the description consists of (dynamic) spatio-temporal relations of the involved space-time histories, before, during and after the interaction (See Table \ref{tbl:interactions} for exemplary interactions). We use $\predTh{occurs-at}(\theta, t)$, and $\predTh{occurs-in}(\theta, \delta)$  to denote that an interaction $\theta$ occurred at a \emph{time point} $t$ or in an \emph{interval} $\delta$, e.g., a person reaching for an object can be defined as follows.

\noindent\begin{minipage}{\columnwidth}
{\footnotesize
\begin{align}
\begin{split}
&\predThF{occurs-in}(\mathsf{reach\_for}(o_i, o_j), \delta) \supset \mathsf{person}(o_i) \wedge\\ 
&\quad \predThF{holds-in}(approaching(body\_part(hand, o_i), o_j), \delta_i) \wedge \\
&\quad \predThF{holds-in}(touches(body\_part(hand, o_i), o_j), \delta_j) \wedge\\
& \quad \mathsf{meets}(\delta_i, \delta_j) \wedge \mathsf{starts}(\delta_i, \delta) \wedge \mathsf{ends}(\delta_j, \delta).\\
\end{split}
\end{align}}
\end{minipage}

These definitions can be used to represent and reason about people interactions involving people movement in the environment, as well as fine-grained activities based on body pose data.

\begin{figure*}[t]
\centering 
\includegraphics[width = .96\textwidth]{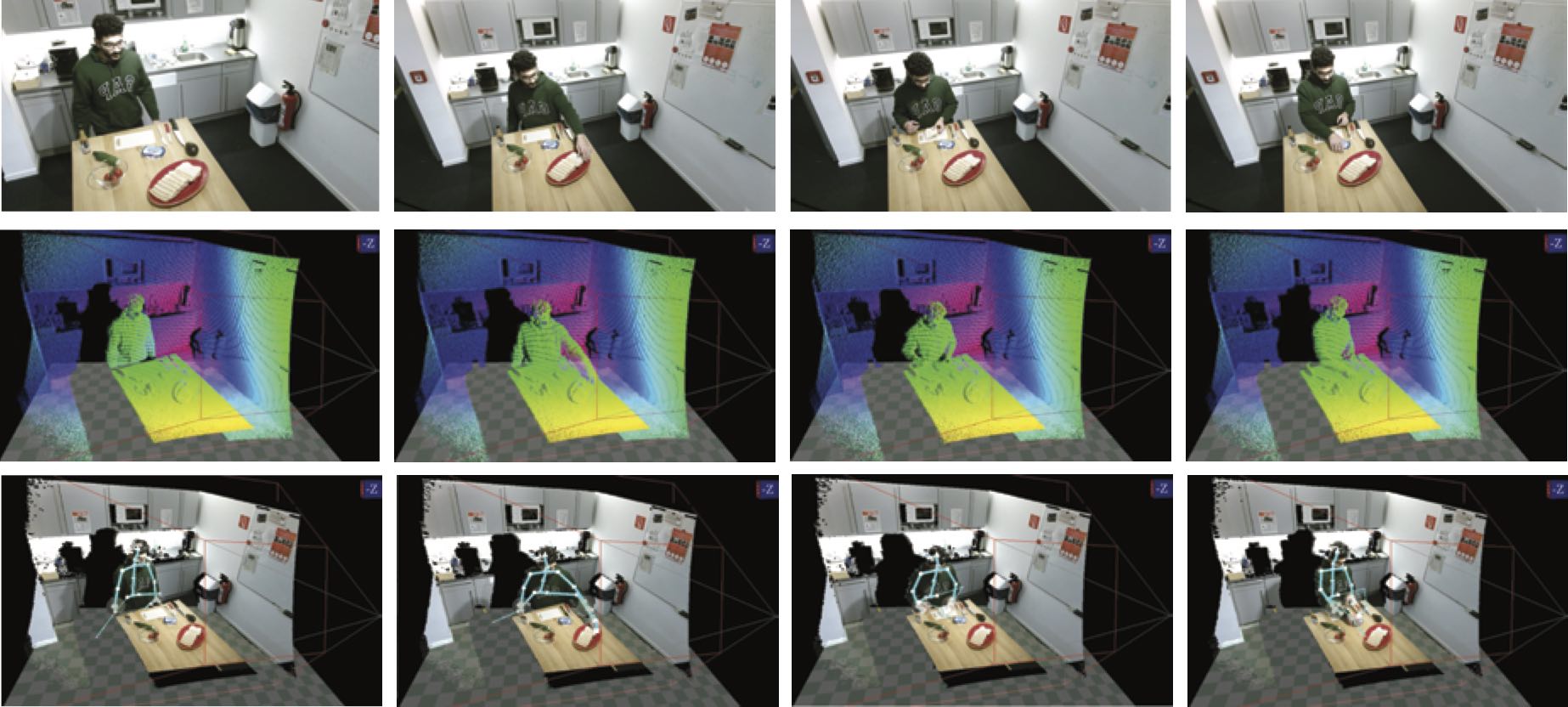}
\caption{{\small\sffamily RGB-D data of Human Activities with Corresponding Skeleton Data}}
\label{fig:rgb-d-data}
\end{figure*}

\section{Application:\\[1pt]Grounding Visuo-Locomotive Interactions}

We demonstrate the above model for grounding everyday activities in perceptual data obtained from RGB-D sensing.
The model has been implemented within (Prolog based) constraint logic programming based on formalisations of qualitative space in CLP(QS) \citep{clpqs-cosit11}. 


\subsubsection*{Human Activity Data}

\textbf{RGB-D Data (video, depth, body skeleton):} We collect data using {\footnotesize\sffamily Microsoft Kinect v2} which provides RGB and depth data.  The RGB stream has a resolution of 1920x1080 pixel at 30 Hz and the depth sensor has a resolution of 512x424 pixels at 30 Hz. Skeleton tracking can track up to 6 persons with 25 joints for each person. Further we use the point-cloud data to detect objects on the table using tabletop object segmentation, based on plane detection to detect the tabletop and clustering of points above the table. For the purpose of this paper simple colour measures are used to distinguish the objects in the scene.

\medskip

\textbf{3D SLAM Data (3d maps, point-clouds, floor-plan structure):}
We collect 3D-SLAM data using Real-Time Appearance-Based Mapping (RTAB-Map) \cite{labbe14online}, which directly integrates with the Robot Operating System (ROS) \cite{ros} for self localisation and mapping under real-time constraints. In particular, for semantic grounding presented in this paper, we use the point-cloud data of the 3D maps obtained from RTAB-Map to extract floor-plan structures by, 1) detection of vertical planes as candidate wall-segments, 2) pre-processing of the wall-segments using clustering and line-fitting, and 3) extraction of room structures based on extracted wall-segments and lines.

\medskip

\noindent\emph{Plane Detection.}\quad Planes in the point-cloud data are detected using a region-growing approach based on the normals of points. To extract candidate wall-segments, we select planes, that are likely to be part of a wall, i.e., horizontal, and sufficiently high or connected to the ceiling. These planes are then abstracted as geometrical entities, specified by their \emph{position}, \emph{size}, and \emph{orientation} (given by the normal), which are used for further analysis.

\medskip

\noindent\emph{Clustering and Line-Fitting.}\quad The detected wall-segments are grouped in a two-stage clustering process using density-based clustering (DBSCAN) \cite{Ester96}, in the first step we cluster wall-segments based on their 2D orientation, in the second step, we align all wall-segments based on the average orientation of the cluster they are in and cluster the wall-segments based on the distance between the parallel lines determined by the wall-segments. We use least square linear regression to fit lines to the resulting wall clusters, which provide the structure of the environment.



\medskip

\noindent\emph{Extracting Room Structures.}\quad Candidate structures for rectangular rooms and corridors are determined by the intersection points of the lines fitted to the wall clusters by considering each intersection point as a possible corner of a room or a corridor. The actual rooms and corridors are then selected based on the corresponding wall segments, projected to the lines.

\begin{figure*}
\begin{center}
  \includegraphics[width=\textwidth]{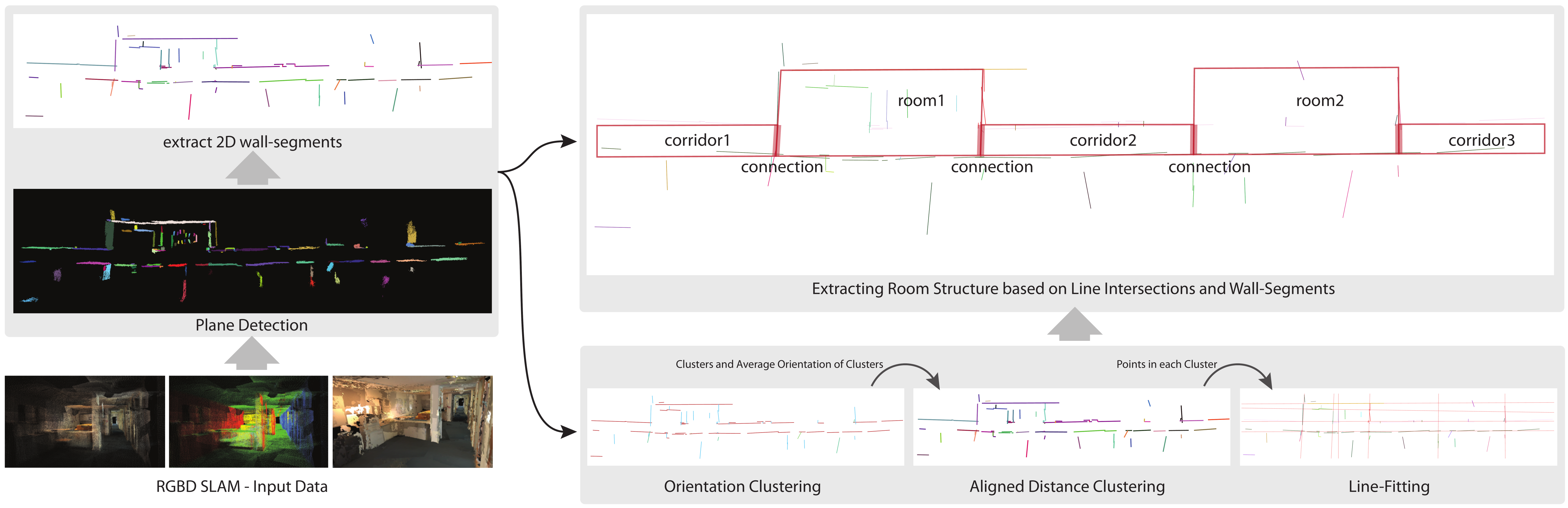}
  \caption{{\small\sffamily Extracting Floor-Plan Semantics from 3D SLAM Data}}
  \label{fig:RGBD-Slam}
\end{center}
\end{figure*}

\subsubsection*{{\color{blue}Ex 1}.\quad Human-Object Interactions}

\textbf{Sample Activity}: ``{\color{blue}Making a Sandwich}''.\quad The activity of making a sandwich is characterised with respect to the interactions between a human and its environment, i.e. objects the human uses in the process of preparing the sandwich.
Each of these interactions is defined by its spatio-temporal characteristics, in terms of changes in the spatial arrangement in the scene (as described in Sec. \ref{sec:activitiy_abstractions}). As an result we obtain a sequence of interactions performed within the track of the particular instance of the activity, grounded in the spatio-temporal dynamics of the scenario. 
As an example consider the sequence depicted in Fig. \ref{fig:rgb-d-data}, the interactions in this sequence can be described as follows:

\medskip

``{\small\sffamily Person1 {\color{blue}\emph{reaches}} for the bread, {\color{blue}\emph{picks up}} a slice of bread, and {\color{blue}\emph{moves}} the hand together with the bread {\color{blue}\emph{back}} to the chopping board.}''

\medskip


The data we obtain from the RGB-D sensor consists of 3D positions of skeleton joints and tabletop objects for each time-point.

\includegraphics[width = 0.95\columnwidth]{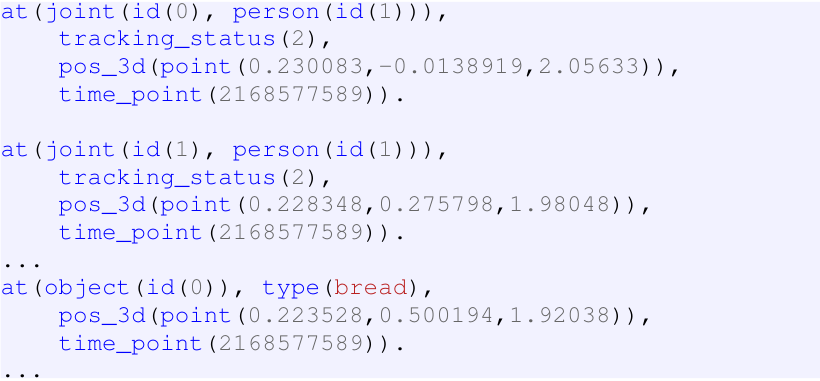}

%
%

%

\medskip

\textbf{Grounded Interaction Sequence}\quad Based on the sensed body-pose data and the detected objects, a sequence of interactions can be queried from the example sequences using the interactive query answering mode of Prolog.

\medskip

\includegraphics[width=0.95\columnwidth]{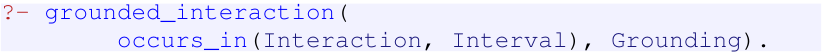}

%
%

This results in all interactions identified in the example sequence and their respective grounding with respect to the spatio-temporal dynamics constituting the interaction,

\medskip

\includegraphics[width=\columnwidth]{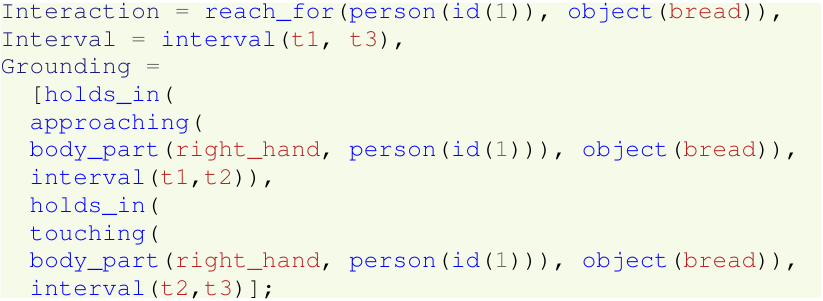}

%


\includegraphics[width=\columnwidth]{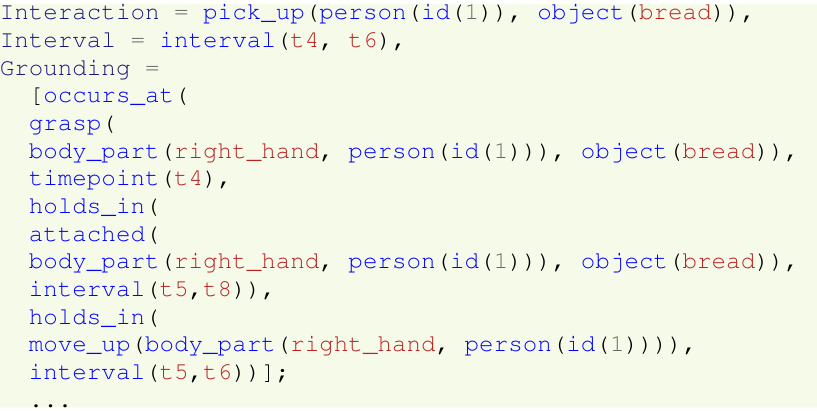}


\medskip

In particular, the interaction $reach\_for(person(id(1)), $ $object(bread))$ occurring between time-point $t1$ and $t3$ is composed of the spatio-temporal pattern of $approaching$, stating that the right hand of person 1 is approaching the bread during time-interval $t1$ to $t2$, and the pattern $touching$, stating that the right hand of person 1 is touching the bread during time-interval $t2$ to $t3$. Similarly the interaction $pick\_up(person(id(1)), object(bread))$ is composed of $grasping$, $attachment$ and $upwards\ movement$, with the difference, that grasping itself is an interaction, that can be further grounded in movement dynamics. 
This kind of declarative grounding can be used, e.g., for relational learning by demonstration, etc. 






\subsubsection*{{\color{blue}Ex 2}.\quad Visuo-Locomotive Interactions}

%




\textbf{Sample Activity}: ``{\color{blue}Indoor Robot Navigation}''.\quad  Robots having to behave and navigate in environments populated by humans have to understand human activities and interactions and have to behave accordingly. In this context, high-level abstractions of human everyday activities and the semantics of the environment can be used to guide robot decision-making to account for humans moving in the environment.

\medskip

As an example consider a robot that has to move from $room1$ to $room2$, during this movement the robot has to pass through the corridor $corridor2$. The structure of the environment is represented as follows:

\medskip

\includegraphics[width=\columnwidth]{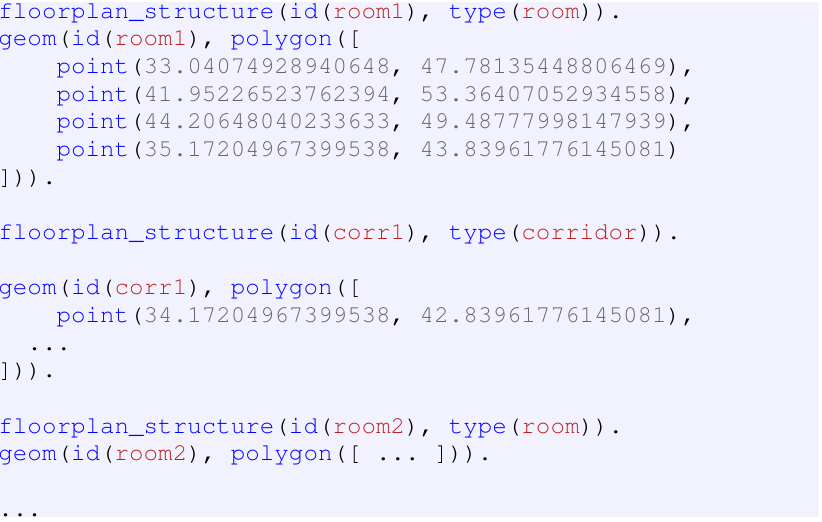}

%
%
%
%
%

\medskip

Based on the extracted semantic floor-plan structure and the detected people in the environment, the robot can make decisions using high-level navigation rules, e.g. defining how to behave in narrow passages, such as the corridor when there are people in the corridor. 
E.g. we can define a rule that a robot can only enter a corridor (the robot can execute the control action $enter(Floorplan\_Structure, T)$), if there is no person in the corridor, or the person is passing the corridor in the same direction as the robot. For this example we use a simple action language for planing the actions of the robot, in this context the predicate $poss\_at(\theta, t)$ defines the spatio-temporal configuration, in which the control action $\theta$ can be executed.


\medskip

%
%

\includegraphics[width=\columnwidth]{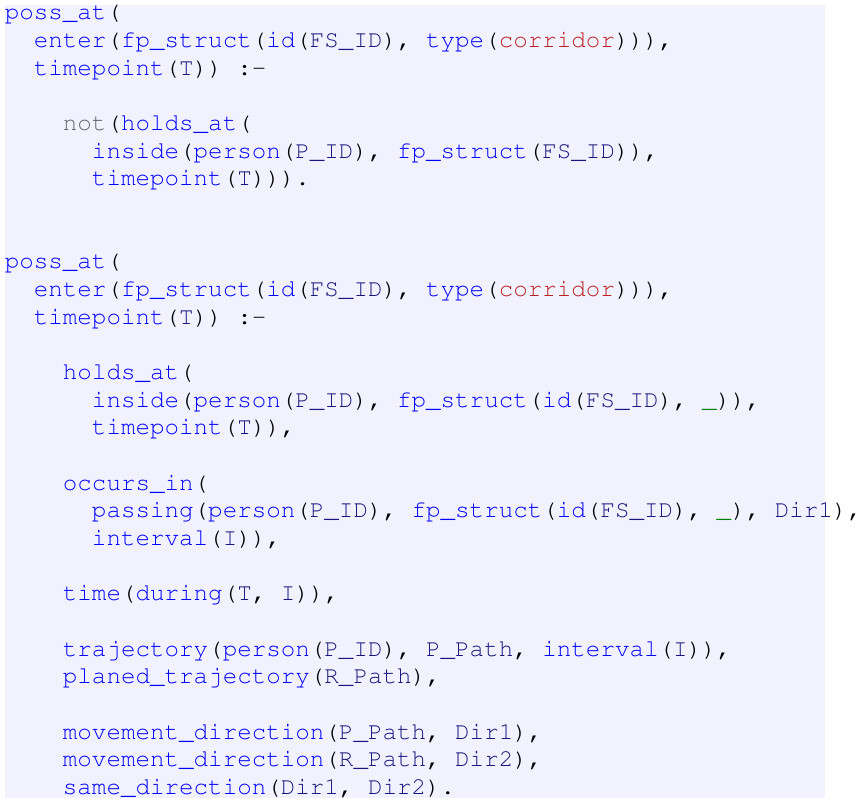}

%
%
%
%
%
%
%
%
%
%

\medskip

The above rules state that the robot can only execute the control action $enter(fp\_struct(id(FS\_ID),$ $type(corridor)))$ if one of the two rules is true.
The first rule simply states that the robot can enter the corridor if there is no person in the corridor. The second rule states, that if there is a person inside the corridor, and the person is passing through the corridor, the robot can enter the corridor, if the trajectory of the person passing through the corridor and the planed path of the robot are passing the corridor in the same direction.

In this way, high-level semantic behaviour descriptions can be used for guiding low-level robot controls, such as path planning, etc.


\section{Summary and Outlook}
Visuo-locomotive sensemaking for practical cognitive robotics in contextualised settings with humans, mobility, and human-machine interaction is a complex endeavour requiring integrative methodologies combining the state of the art from several research areas such as vision, AI, HCI, and engineering. Our research emphasis the particular merit of combining visual processing with semantic knowledge representation and reasoning techniques rooted in artificial intelligence, particularly commonsense reasoning. We have presented a declarative commonsense model for grounding embodied visuo-locomotive interactions; the proposed model ---consisting of a formal characterisation of {space, time, space-time, and motion patterns}--- is geared towards knowledge representation and reasoning capabilities such as  commonsense abstraction, learning, reasoning, embodied simulation. With a particular focus on the representation of space-time histories and motion patterns, the model is illustrated with select RGB-D datasets corresponding to representative activities from a larger dataset of everyday activities. Immediate next steps involve integration with state of the art robot control platforms such as ROS; this will be accomplished via integration into the {\small\sffamily{ExpCog}} commonsense cognition robotics platform for experimental / simulation purposes, and within {\small\sffamily{openEASE}} as a state of the art cognition-enabled control of robotic control platform for real robots.\footnote{{\sffamily{ExpCog}} -- \url{http://www.commonsenserobotics.org}\\{\small\sffamily{openEASE}} -- \url{http://ease.informatik.uni-bremen.de/openease/}}


\section*{Acknowledgements}
{\small\sffamily

This research is partially funded by the Germany Research Foundation (DFG) via the Collabortive Research Center (CRC) EASE -- Everyday Activity Science and Engineer\-ing (\url{http://ease-crc.org}). We also acknowledge the support of Omar Moussa, Thomas Hudkovic, and Vijayanta Jain in preparation of parts of the overall activity dataset.}
%

{\small
\bibliographystyle{ieee}


}

\end{document}